\documentclass{article}
\usepackage{PRIMEarxiv}
\usepackage{inputenc} 
\usepackage[T1]{fontenc}    
\usepackage{hyperref}       
\usepackage{url}            
\usepackage{booktabs}       
\usepackage{amsfonts}       
\usepackage{nicefrac}       
\usepackage{microtype}      
\usepackage{lipsum}
\usepackage{graphicx}
\graphicspath{{media/}}     

\usepackage{algorithm}
\usepackage{algorithmicx}
\usepackage{algpseudocode}
\usepackage{amsmath}
\usepackage{subcaption}
\usepackage{hyperref}

\title{DKDL-Net: A Lightweight Bearing Fault Detection Model via Decoupled Knowledge Distillation and LoRA Fine-tuning
}

\author{
  Ovanes Petrosian \thanks{These authors contributed equally to this work.} \\
  SPBU\\
  \and
  \textbf{Li Pengyi}\footnotemark[1] \\
  SPBU\\
  st112719@student.spbu.ru\\
  \and
  \textbf{He Yulong} \\
  SPBU
  \and
  \textbf{Liu Jiarui} \\
  SPBU
  \and
  \textbf{Sun Zhaoruikun} \\
  SPBU
  \and
  \textbf{Fu Guofeng} \\
  SPBU
  \and
  \textbf{Meng Liping} \\
  SPBU
}

\begin{document}

\maketitle
\begin{abstract}
Rolling bearing fault detection has developed rapidly in the field of fault diagnosis technology, and it occupies a very important position in this field. Deep learning-based bearing fault diagnosis models have achieved significant success. At the same time, with the continuous improvement of new signal processing technologies such as Fourier transform, wavelet transform and empirical mode decomposition, the fault diagnosis technology of rolling bearings has also been greatly developed, and it can be said that it has entered a new research stage. However, most of the existing methods are limited to varying degrees in the industrial field. The main ones are fast feature extraction and computational complexity. The key to this paper is to propose a lightweight bearing fault diagnosis model DKDL-Net to solve these challenges. The model is trained on the CWRU data set by decoupling knowledge distillation and low rank adaptive fine tuning. Specifically, we built and trained a teacher model based on a 6-layer neural network with 69,626 trainable parameters, and on this basis, using decoupling knowledge distillation (DKD) and Low-Rank adaptive (LoRA) fine-tuning, we trained the student sag model DKDL-Net, which has only 6838 parameters. Experiments show that DKDL-Net achieves 99.48\% accuracy in computational complexity on the test set while maintaining model performance, which is 0.58\% higher than the state-of-the-art (SOTA) model, and our model has lower parameters. Our code is available at Github link: https://github.com/SPBU-LiPengyi/DKDL-Net.git.
\end{abstract}

\keywords{Bearing fault diagnosis \and Convolutional neural network \and Model compression \and Low-Rank Adaptation}

\section{Introduction}\label{introduction}
Rolling bearings are common components in rotating machinery, designed to reduce friction during rotation and thereby enhance the safety of the equipment. Studies in the industrial sector indicate that approximately 40\%-70\% of mechanical failures are caused by bearing faults \cite{fault1, fault2, fault3, fault4}. However, traditional bearing fault detection methods are time-consuming. In the current era of artificial intelligence, it is crucial to use deep learning techniques for fault detection tasks. Training an efficient and accurate bearing fault detection model can not only improve detection efficiency but also significantly reduce economic losses \cite{loss1}.

Bearing fault diagnosis is typically based on acoustic signals and bearing vibration signals \cite{bearingway1}. Bearing faults cause abnormal vibrations and sounds, making it possible to detect faults by diagnosing these anomalies. In practical detection tasks, sensors are usually installed on machine tools to capture the bearing's sound signals.

There are various approaches to vibration fault detection based on sound signals and bearings, including methods using deep belief networks (DBN) \cite{bdbn, bdbn1, bdbn2}, convolutional neural network (CNN) \cite{cnn1, cnn2, cnn3}, deep autoencoders (DAE) \cite{dae1, dae2, dae3}, generative adversarial networks (GAN) \cite{gan1, gan2, gan3}, and deep transfer learning (DTL) \cite{dtl1, dtl2, dtl3, dtl4}. What's common among these models is that they differ only in the way neural networks are constructed and training strategies, while the data type remains the same. Additionally, they all require a large amount of data.

Over the past 5 years, bearing fault detection tasks have commonly used models such as Convolutional Neural Networks (CNN) \cite{cnn1, cnn2, cnn3}, Recurrent Neural Networks (RNN) \cite{rnn1, rnn2, rnn3}, and Long Short-Term Memory Networks (LSTM) \cite{lstm1, lstm2, lstm3, mcnn-lstm}. Experiments have shown that lightweight models fail to achieve the desired results, with the number of model parameters affecting the final outcome. For instance, LEFE-Net \cite{lefe-net}, WDCNN \cite{wdcnn}, and MCNN-LSTM \cite{mcnn-lstm} models can achieve an accuracy of over 98.50\% on the CWRU \cite{cwru1, cwru2} dataset, but their trainable parameters exceed 50,000. In contrast, the CLFormer \cite{clformer} model has 4,980 parameters but an accuracy of less than 95\%.

The key to whether a model can be applied in industry lies in its lightweight nature, high accuracy, and robustness. Models based on acoustic signals typically transform time-domain signals into frequency-domain signals using techniques such as Short-Time Fourier Transform (STFT) \cite{stft1}, Empirical Mode Decomposition (EMD) \cite{emd1, emd2}, and envelope spectrum analysis. Neural network models are then built to extract features, and finally, classification is performed through fully connected layers to achieve bearing fault diagnosis. Based on this process, we have constructed a lightweight bearing fault detection model.

In this paper, first, we trained a Teacher model with a large number of parameters, and then trained a lightweight Student model using DKD \cite{dkd}. This Student model has only one convolutional layer, one pooling layer, and one fully connected layer. The Student model, guided by the Teacher model during training, has a significantly lower number of parameters. However, we found that the accuracy of the Student model was 2\% lower than that of the Teacher model. To address this, we introduced a Low-Rank Adaptation (LoRA) \cite{lora} to fine-tune the Student model, which improved accuracy by 1.5\% with a relatively short training time. Compared to traditional knowledge distillation and fine-tuning methods, the combination of DKD and LoRA fine-tuning ensures model performance while significantly reducing the number of training parameters.

We propose a model for industrial application in rolling bearing fault detection and contributions are fourfold:

\begin{enumerate}
\item We developed a lightweight rolling bearing fault detection model based on DKD, characterized by a single-layer neural network, It is compressed by 90.20\% compared to the teacher model (6-layer neural network).
\item We improved the model's performance after knowledge distillation by using a LoRA fine-tuning method, addressing the performance degradation issue.
\item Compared to other lightweight models, our approach demonstrates superior performance on the CWRU dataset, with average accuracy, precision, recall, and F1-Score all higher than those of other models. Our F1-Score reached 99.50\%, and the trainable parameters of our model are only 6,838.
\item Compared to the SOTA model our model improved the F1-Score by 0.58\%.
\end{enumerate}

\section{Related Work}\label{related work}

\begin{figure}[!hbtp]
    \centering
    \includegraphics[width=0.6\linewidth]{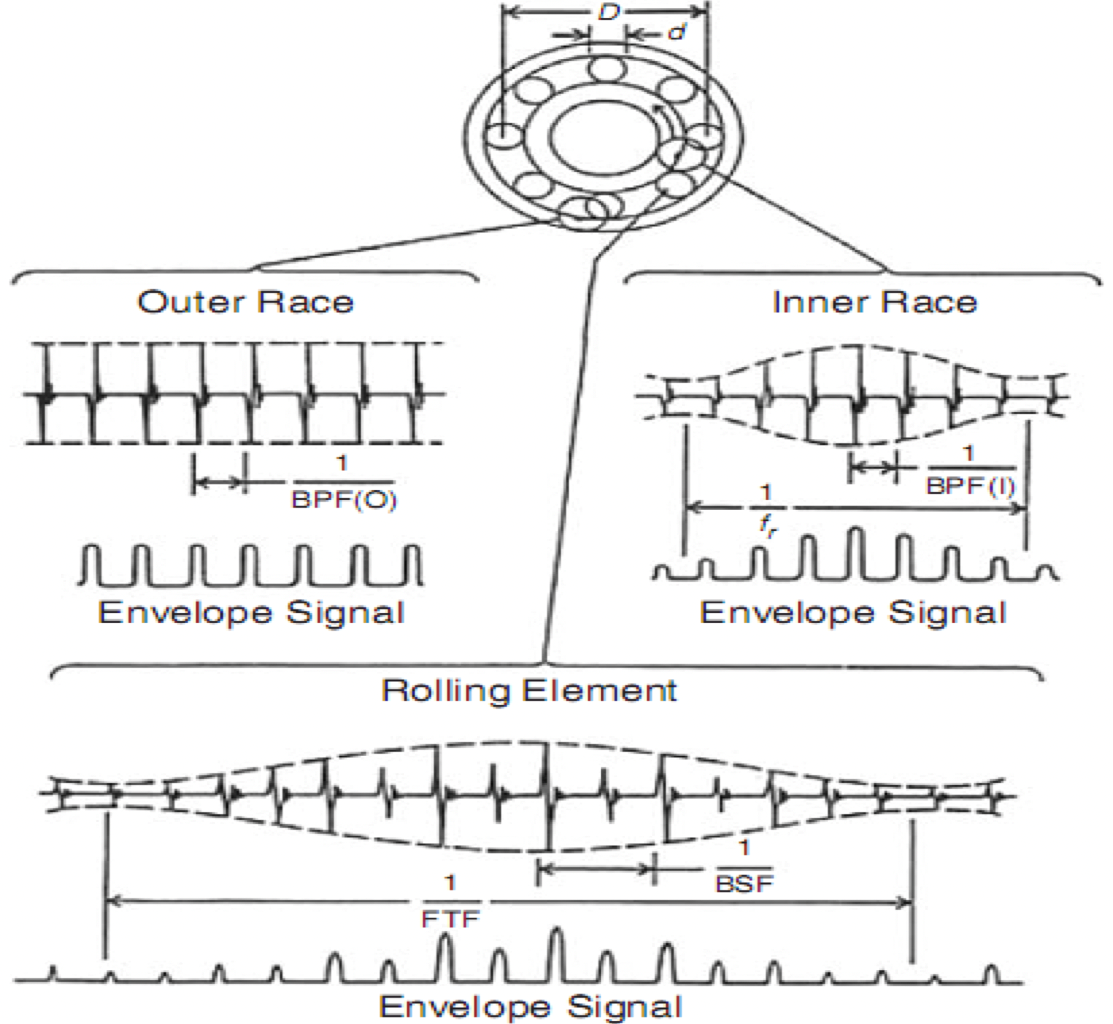}
    \caption{Structure of Rolling Bearings}
    \label{bearing}
\end{figure}

\textbf{Bearing fault detection.} The fault detection of rolling bearings typically involves checking whether the three components of the bearing (outer race, inner race, rolling elements) are faulty. The structure of the bearing is visualized as shown in Fig. \ref{bearing}. There are 10 categories of bearing sounds \cite{neupane2020bearing}, where one category represents health components, three categories represent damaged rolling elements, three categories represent damaged inner races (IR), and three categories represent damaged outer races (OR). Once abnormal vibrations occur in the bearing, it indicates a fault.

\textbf{Convolutional Neural Network.} There has been extensive work on bearing fault detection models based on convolutional neural networks. The Adaptive Deep Convolutional Neural Network (ADCNN) \cite{islam2019automated} utilizes the distribution of key information in discrete frequency bands to diagnose the health status of rotating components. The 2D LeNet-5 \cite{wan2020rolling} is an improved bearing fault detection model based on LeNet, which performs one-dimensional convolution and pooling operations directly on the raw vibration signals without any preprocessing. The lightweight multi-scale CNN \cite{shi2020enhanced} introduces depthwise separable convolution in a multi-scale CNN to reduce the model's storage and computational costs. The hybrid deep neural network model integrated with Principal Component Analysis (PCA) \cite{you2022rolling} based model combines RNN and CNN, using bidirectional long short-term memory (BiLSTM) to enhance the extraction of time-series data features, and finally employs an Attention mechanism to improve the model's accuracy. KDSCNN \cite{ji2022neural} is a CNN model based on KD and model parameters are only 5890. BearingPGA-Net \cite{bearingpga} is a lightweight model trained with DKD and applied on an FPGA module. MCNN-LSTM \cite{mcnn-lstm} is a model based on CNN and LSTM, which first uses CNN to extract features and then processes classification according to the time series information from LSTM. FaultNet \cite{magar2021faultnet} and WDCNN \cite{wdcnn} are CNN models based on convolutional kernels of different scales. 

While these models have achieved commendable performance, none of them offers a low-parameter model that combines high efficiency and accuracy.

\textbf{Model compression.} Neural network compression involves various techniques, including model pruning, parameter quantization, low-rank decomposition, knowledge distillation (KD), and lightweight model design \cite{li2023model}. Model pruning \cite{liang2021pruning} can reduce the accuracy and lead to irregular structures. Parameter quantization \cite{liang2021pruning} requires extensive training and fine-tuning. Low-rank decomposition \cite{nguyen2019low} involves decomposing and training layer-by-layer parameters, which can be challenging for highly complex models. KD \cite{gou2021knowledge} is a method where a teacher model guides the training of a student model. The teacher model is typically complex, while the student model can have a simple structure, such as a single-layer network. This method achieves significant compression with minimal accuracy loss. Lightweight model design is challenging to combine with other compression methods \cite{choudhary2020comprehensive}. If improperly combined, it can lead to lower performance and poor generalization. 

All in all, KD, on the other hand, can significantly reduce model parameters while maintaining high accuracy. In fact, the algorithm mentioned in this article shows an accuracy drop of less than 2\%.

DKD \cite{dkd} is an algorithm that addresses the inefficiency problem of traditional knowledge distillation. It enhances the knowledge distillation process by decomposing and separately optimizing different components of the knowledge transfer from the teacher model to the student model. DKD divides traditional knowledge distillation into two independent parts: Target Class Knowledge Distillation (TCKD) and Non-Target Class Knowledge Distillation (NCKD), and defines separate loss functions to optimize the model. DKD is already a mature algorithm and has been widely applied in various tasks such as large language models LLMs \cite{zhu2023survey}, LVM \cite{wang2023review}, and object detection \cite{li2023object}. It is highly efficient for model compression.

\textbf{Fine-tuning.} LoRA was first proposed for fine-tuning LLMs \cite{lora} for downstream tasks, such as fine-tuning GPT for grammar correction \cite{loragrammargpt}, fine-tuning Llama for fire safety training \cite{lorasafety}, and fine-tuning Stable Diffusion to enhance image generation for specific tasks \cite{lorastable}. LoRA achieves this by decomposing the weight matrices into low-rank matrices. This approach significantly reduces the number of parameters that need to be fine-tuned, decreases the storage requirements of the model, and lowers the computational complexity. As a result, both inference and training become more efficient.

In CNNs, LoRA can be used to compress the model by applying low-rank decomposition to the convolutional kernels \cite{loracnn1}, thereby reducing the number of parameters to be trained. This approach reduces the model's storage and computational costs. After fine-tuning a CNN model with LoRA, both inference speed and training time are greatly improved. Using low-rank decomposition for object detection \cite{loracnn} enhances the model's performance while reducing its parameters. Compared to the original CNN model, a LoRA fine-tuned model can operate under lower hardware resource conditions.

Our idea is to propose a model based on DKD training and LoRA fine-tuning. And model is not only fast in reasoning but also possesses high accuracy.

\section{Method}\label{method}

\subsection{CNN and Classification}
A CNN consists of $n$ ($n \in \{0, 1, 2, \ldots, n\}$) convolutional layers and $m$ ($m \in \{0, 1, 2, \ldots, m\}$) pooling layers. Convolutional layers use multiple filters to convolve over feature maps, extracting features. The extracted features are then passed through activation functions for non-linear transformations, helping to mitigate issues like vanishing gradients during training. Pooling layers are used to reduce the number of features, thus decreasing computational complexity.

The output value $a_j^l$ of the $j-th$ unit of the convolutional layer $l$ is given by Eq. \ref{cnnlayer}.
\begin{equation}\label{cnnlayer}
    a_j^l=f(b_j^l+\sum_{i\in M_j^l} a_i^{l-1}*k_{ij}^l)
\end{equation}
The activation value $a_j^l$ in the pooling layer $l$ is given by Eq. \ref{poolinglayer}.
\begin{equation} \label{poolinglayer}
    a_j^l=f(b_j^l+\beta_j^l down(a_j^{l-1},M^l))
\end{equation}
where $down(*)$ represents the pooling function. Common pooling functions include average pooling, max pooling, min pooling, and stochastic pooling. \(b_j^l\) denotes the bias, \(\beta_j^l\) represents the multiplier residual, and \(M^l\) is the size of the pooling window used in the l-th layer.

After the model extracts features through convolutional layers, it needs to classify the feature data. Generally, classification tasks uses a cross-entropy (CE) \cite{celoss} loss as a objective function, the expression for CE is as in Eq. \ref{celoss}. By optimizing the cross-entropy loss, the model's classification accuracy is improved.
\begin{equation}\label{celoss}
    L_{CE} = \sum^N_{n=1}-y_{n}log(p_n),\ p_n = \frac{e^{z_n}}{\sum^{N}_{i=1}e^{z_i}}
\end{equation}
where $N$ denotes the number of categories, $y_n$ denotes true label, $log(p_n)$ denotes natural logarithm of the predict probability of n-th label from model, $z_n$ is raw scores from model output of n-th label, $p_n$ predicted probability for the n-th label.

\subsection{DKD}

Although there are various types of knowledge distillation, we use DKD for improvement our model. First, in the classical KD, the $logit$, $l_i$, computed for each class is converted to a probability $p_i$ by using the $softmax(*)$ function as shown in Eq. \ref{classical}.
\begin{equation}\label{classical}
    p_{i}=\frac{\exp \left(l_{i}\right)}{\sum_{j=1}^{N} \exp \left(l_{j}\right)}
\end{equation}
where $N$ denotes the number of classes.
\begin{equation}
    b = [p_t, p_{\neg t}] \in \mathbb{R}^{1 \times 2}
\end{equation}
Then, we use binary probabilities $ b = [p_t, p_{\neg t}] \in \mathbb{R}^{1 \times 2} $ to distinguish between predictions related and unrelated to the target class, where $p_t$ denotes the target class and $p_{\neg t}$ denotes the non-target class, calculated as shown Eq. \ref{bnp}.
\begin{equation} \label{bnp}
    p_{t} = \frac{\exp(l_t)}{\sum_{j=1}^N \exp(l_j)},
p_{-t} = \frac{\sum_{d=1, d \neq t}^N \exp(l_d)}{\sum_{j=1}^N \exp(l_j)}.
\end{equation}
Meanwhile, we use $\Tilde{p}_i = \frac{p_i}{p_t}$ to denote the probability between non-target categories (i.e., without considering the target category t) calculated as Eq. \ref{NT}.
\begin{equation} \label{NT}
    \tilde{p}_i = \frac{\exp(l_i)}{\sum_{j=1, j \neq i}^{N} \exp(l_j)}
\end{equation}
Classical KD uses KL-Divergence as the loss function, and further, we re-represent KD using the binary probability $b$ and the non-target class $\Tilde{p}$, T and S stand for teacher and student, respectively. represented as Eq. \ref{KD}. 
\begin{equation} \label{KD}
    \begin{alignedat}{2}
        \text{KD} &= \text{KL}(\mathbf{p}^T \parallel \mathbf{p}^S) = p^T_t \log\left(\frac{p^T_t}{p^S_t}\right) + \sum_{i=1, i \neq t}^N p^T_i \log\left(\frac{p^T_i}{p^S_i}\right)
    \end{alignedat}
\end{equation}
Simplifying, we can rewrite Eq. \ref{KD} as Eq. \ref{reKD} X by Eq. \ref{classical} and Eq. \ref{NT}.
\begin{equation}\label{reKD}
    \begin{aligned}
        \text{KD} = & p^T_t \log\left(\frac{p^T_t}{p^S_t}\right) + p^T_{\neg t} \sum_{i=1, i\neq t}^{D} \tilde{p}_i^{T} \left(\log\left(\frac{\tilde{p}_i^{T}}{\tilde{p}_i^{S}}\right) \right. + \left. \log\left(\frac{p^T_{-t}}{p^S_{-t}}\right)\right) \\
        = & p^T_t \log\left(\frac{p^T_t}{p^S_t}\right) + p^T_{\neg t} \log\left(\frac{p^T_{-t}}{p^S_{-t}}\right) + p^T_{\neg t} \sum_{i=1, i\neq t}^{D} \tilde{p}_i^{T} \log\left(\frac{\tilde{p}_i^{T}}{\tilde{p}_i^{S}}\right)
    \end{aligned}
\end{equation}
Simplifying, KD can then be rewrite as Eq. \ref{rewKD}.
\begin{equation}\label{rewKD}
    \text{KD} = \text{KL}(b^T \parallel b^S) + (1 - p^T_t) \text{KL}(\Tilde{p}^T \parallel \Tilde{p}^S)
\end{equation}
where $\text{KL}(b^T \parallel b^S)$ denotes the similarity between teacher and student probabilities in the target class, which we refer to as Target Class Knowledge Distillation (TCKD), and $\text{KL}(\Tilde{p}^T \parallel \Tilde{p}^S)$ denotes the similarity between teacher and student probabilities in the non-target class, which we refer to as Non-Target Class Knowledge Distillation (NCKD), and thus we can rewrite KD Eq. \ref{endKD}.
\begin{equation} \label{endKD}
    \text{KD} = \text{TCKD} + (1 - p^T_t) \text{NCKD}
\end{equation}
Observing the latest KD formulation, we find that on the one hand NCKD is coupled with $(1 - p^T_t)$, which would suppress NCKD for well-predicted samples. On the other hand, the weights of NCKD and TCKD are coupled in the classical KD framework, which does not allow to change the weights of each term in order to balance the importance. Therefore DKD introduces two hyperparameters $\alpha$ and $\beta$ as weights for TCKD and NCKD, respectively. Thus the loss function of DKD can be written as Eq. \ref{DKD}.
\begin{equation} \label{DKD}
    \mathcal{L}_{DKD} = \alpha * \text{TCKD} + \beta * \text{NCKD}
\end{equation}
where $\beta$ replaces $(1 - p^T_t)$ to prevent inhibiting the effectiveness of the NCKD, and secondly, $\alpha$ and $\beta$ can be allowed to be adjusted to achieve a balance of importance. By optimizing this decoupling loss, the knowledge gained by the teacher model is more easily transferred to the student model, thus improving the performance of the student network.

\subsection{LoRA}
LoRA is a method for efficiently fine-tuning pre-trained models on specific tasks. This algorithm reduces the fine-tuning parameters using a low-rank approach while enhancing the model's performance on the given task. Fine-tuning with fewer parameters can achieve over 90\% of the performance of full fine-tuning. For a pre-trained model with a weight parameter matrix \(W_0 \in \mathbb{R}^{d \times k}\), \(\Delta W \in \mathbb{R}^{d \times k}\) represents the fine-tuning parameters for a specific task. \(\Delta W\) is a lower-dimensional parameter matrix that can be expressed as \(B \times A\), where \(B \in \mathbb{R}^{d \times r}\) and \(A \in \mathbb{R}^{r \times k}\), with \(r \ll k\). Thus, the parameter count of \(\delta W\) is smaller than that of \(W_0\). 

The LoRA algorithm as show in the Eq. \ref{lora}, with its key idea being to decompose the parameter matrix using a low-rank matrix decomposition.
\begin{equation} \label{lora}
    h=W_0x+\Delta Wx=W_0 x+BAx
\end{equation}
\subsection{DKDL-Net Model}
\begin{figure*}[!hbtp]
    \centering
    \includegraphics[width=1\linewidth]{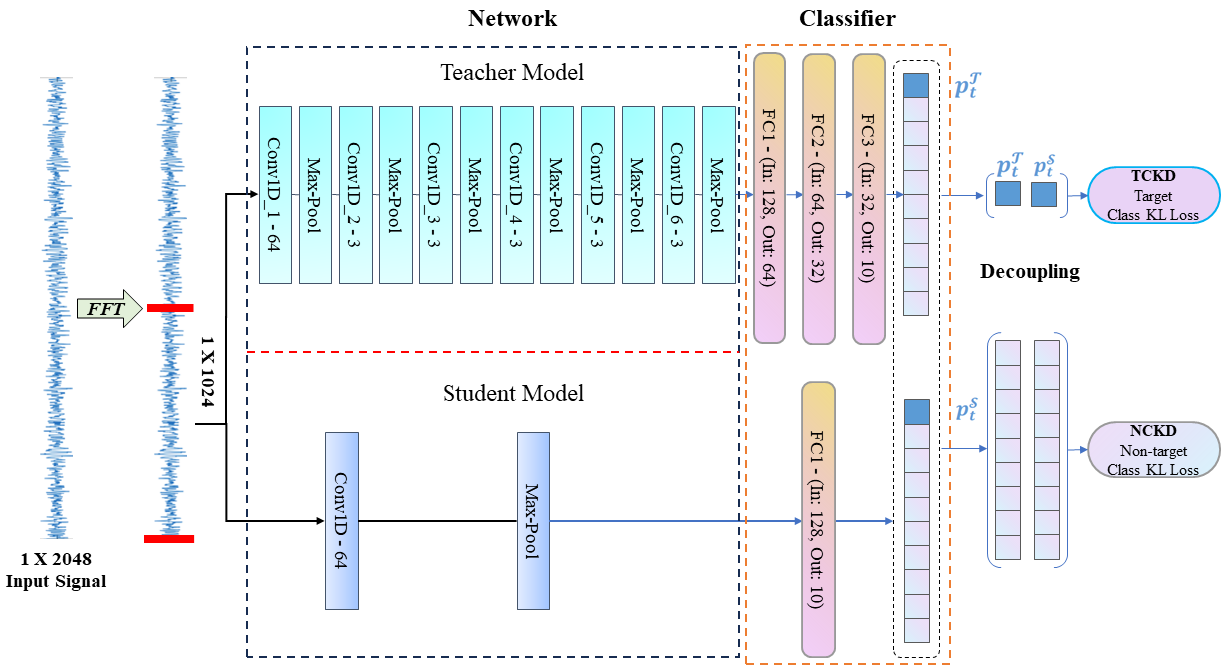}
    \caption{Architecture of the Model. (top) Teacher Model, (below) Student Model}
    \label{st model}
\end{figure*}

The DKDL-Net model is based on the DKD approach, where a Teacher model guides the training of a Student model. The model framework is illustrated in Fig. \ref{st model}. The Teacher model is a large-scale model with a substantial number of parameters, and its increased depth enhances accuracy in bearing fault detection. However, the large parameter size of the Teacher model results in slow inference speed, making it unsuitable for efficient industrial tasks. Therefore, we trained the Student model using the DKD method. This model is a single-layer neural network, meaning it has fewer parameters and faster inference speed. However, since the Student model is derived from significant parameter compression, its accuracy decreases. In simple experimental analyses of bearing fault detection, the Student model's accuracy is approximately 2\% lower compared to the Teacher model.

\begin{figure*}[!hbtp]
    \centering
    \includegraphics[width=1\linewidth]{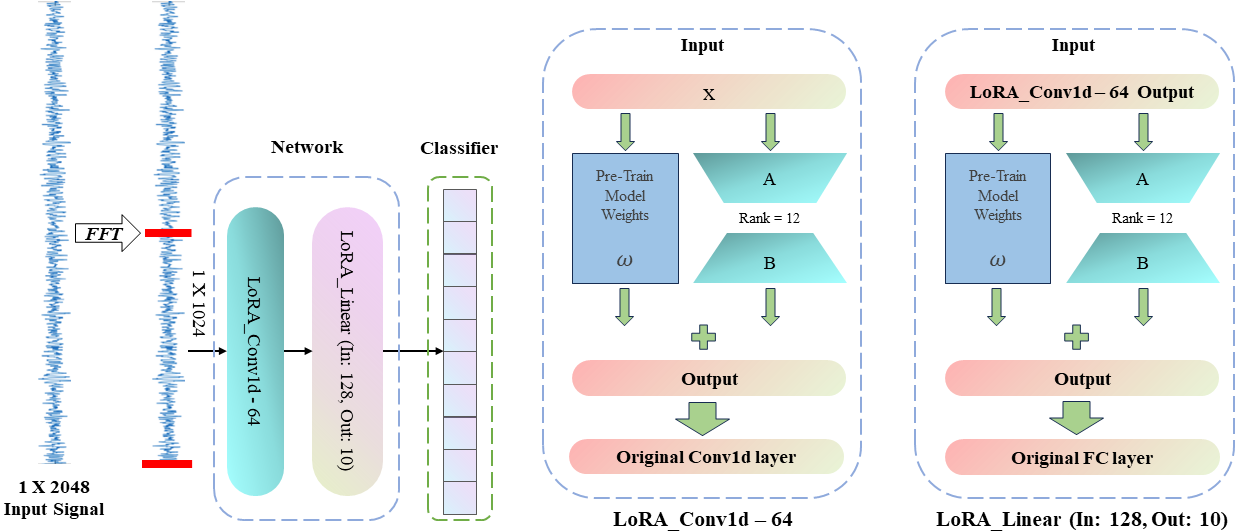}
    \caption{Architecture of the DKDL-Net Model}
    \label{DKD model}
\end{figure*}

Through extensive research, we found that we can further fine-tune the Student model using the LoRA approach. The model framework is illustrated in Fig. \ref{DKD model}. Typically, LoRA involves low-rank decomposition of the model's convolutional and fully connected layers. In this task, we cannot reduce the model's parameters further, as experiments have shown that this would decrease the model's accuracy. Therefore, we integrated the LoRA module into the single-layer network.

In the DKDL-Net network, we copied the parameters from the Student model. The parameters of the $A$ matrix in the LoRA module are initialized with data following a normal distribution $N(0, \sigma^2)$, while the $B$ matrix is initialized to 0. This approach allows us to enhance the model's accuracy by adding only a small number of parameters.

Regarding the training of the model, we utilize CE Loss as the objective function for training the Teacher model, defined mathematically as Eq. \ref{celoss}. For training the Student model, we employ a combination of TCKD Loss, NCKD Loss, and CE Loss as the loss functions. It is essential to balance CE Loss and DKD Loss during the training process, where DKD Loss is the sum of TCKD Loss, CE Loss and NCKD Loss, defined mathematically as Eq. \ref{trainLoss}.
\begin{equation} \label{trainLoss}
    \mathcal{L} = (1 - \gamma) \underbrace{L_{CE}}_{\text{CE loss}} + \gamma \underbrace {\left( \alpha \ast \underbrace{\text{TCKD}}_{\text{TCKD Loss}} + \beta \ast \underbrace{\text{NCKD}}_{\text{NCKD Loss}} \right)}_{\text{DKD Loss}}
\end{equation}
where $\gamma$ is a learnable parameter to balance CE and DKD Loss.

Finally, incorporating the LoRA plug-and-play module into the Student model, we fine-tune it using the CWRU dataset with CE Loss as the loss function, mathematically defined as Eq. \ref{celoss}.

\begin{algorithm}
\caption{The training phase of DKDL-Net model}\label{algo1}
\begin{algorithmic}[1]
\Require Model parameters $W_{conv1d}^{freez}$, $W_{fc}^{freez}$, $B_{conv1d}^{freez}$, $B_{fc}^{freez}$ after training with DKD.
\Statex \textbf{Input:} 
\begin{itemize}
    \item CWRU dataset ($x_1^{CWRU}$, $x_2^{CWRU}$, ..., $x_n^{CWRU}$)
    \item ($x_1^{CWRU}$, $x_2^{CWRU}$, ..., $x_n^{CWRU}$ ) of input to FFT get ($x_1^{f}$,...,$x_n^{f}$)
    \item Use chunk get ($x_1^{m}$,$x_2^{m}$,...,$x_n^{m}$)
\end{itemize}
\Statex \textbf{Initialisation:} 
\begin{itemize}
    \item initial $A_{conv1d}$ , $B_{conv1d}$ , matrix ,Rank = r
    \item initial $A_{fc}$ , $B_{fc}$ , matrix ,Rank = r
\end{itemize}

\For{$Epoch^{th}$ training iteration in total epoch, $E$}
    \State $\hat{F}^L_{\text{$conv1d$}}$ = $B_{conv1d}$ $\cdot$ $A_{conv1d}$ $\cdot$ ($x_1^{m,e}$, $x_2^{m,e}$, ..., $x_n^{m,e}$)
    \State $\hat{F}^{initial}_{\text{$conv1d$}}$ = $w_{conv1d}^{freez}$ $\cdot$ $\hat{F}^L_{\text{$conv1d$}}$ $+$ $B_{conv1d}^{freez}$
    \State $\hat{F}^L_{\text{$fc$}}$ = $B_{fc}$ $\cdot$ $A_{fc}$ $\cdot$ $\hat{F}^{initial}_{\text{$conv1d$}}$
    \State $\hat{F}^{initial}_{\text{$fc$}}$ = $w_{fc}^{freez}$ $\cdot$ $\hat{F}^L_{\text{$fc$}}$ $+$ $B_{fc}^{freez}$
    \State $X_{output}^{e}$ = $\hat{F}^{initial}_{\text{$fc$}}$
    \State Calculate loss function and gradient, then use Adam optimizer, update parameters by Backpropagation algorithm
\EndFor
\Statex \textbf{Output:} DKDL-Net model weights file (.pth)
\end{algorithmic}
\end{algorithm}

\subsection{Structural configuration of the model}

The framework of the Teacher model is shown in Fig. \ref{st model} (top), and the relevant parameter configurations for the model's input, output, and convolutional kernel size are presented in Table \ref{Teacher Model}.

\begin{table*}[!hbtp]
    \centering
    \caption{Teacher model parameters applied}\label{Teacher Model}
    \begin{tabular*}{\textwidth}{@{\extracolsep\fill}llllll}
    \toprule
    Name & Kernel size/stride & Input size & Output size & Activation function & \#Parameters \\ 
    \midrule
    Conv1D\_1 & (64, ) / 8 & 1 $\times$ 1024 & 16 $\times$ 128 & ReLU & 1072 \\
    Pooling\_1 & 2 / 2 & 16 $\times$ 128 & 16 $\times$ 64 &  & 0 \\
    Conv1D\_2 & (3, ) / 1 & 16 $\times$ 64 & 32 $\times$ 64 & ReLU & 1632 \\
    Pooling\_2 & 2 / 2 & 32 $\times$ 64 & 32 $\times$ 32 &  & 0 \\
    
    Conv1D\_3 & (3, ) / 1 & 32 $\times$ 32 & 64 $\times$ 32 & ReLU & 6336 \\
    Pooling\_3 & 2 / 2 & 64 $\times 32$ & $64 \times 16$ &  & 0 \\
    Conv1D\_4 & (3, ) / 1 & 64 $\times$ 16 & 64 $\times$ 16 & ReLU & 12480 \\
    Pooling\_4 & 2 / 2 & $64 \times 16$ & $64 \times 8$ &  & 0 \\
    Conv1D\_5 & (3, ) / 1 & 64 $\times$ 8 & 64 $\times$ 8 & ReLU & 12480 \\
    Pooling\_5 & 2 / 2 & $64 \times 8$ & $64 \times 4$ &  & 0 \\
    Conv1D\_6 & (3, ) / 1 & 64 $\times$ 4 & 128 $\times$ 2 & ReLU & 24960 \\
    Pooling\_6 & 2 / 2 & $128 \times 2$ & $128 \times 1$ &  & 0 \\
    FC\_1 &  & 128 & 64 & ReLU & 8256 \\
    FC\_2 &  & 64 & 32 &  & 2080 \\
    FC\_3 &  & 32 & 10 &  & 330 \\
    \multicolumn{5}{c}{Total of trainable parameters} & 69626 \\
    \hline
    \end{tabular*}
\end{table*}

The Student model is a single layer network and framework shown in Fig. \ref{st model} (below), and the relevant parameter configurations for the model's input, output, and convolutional kernel size are presented in Table \ref{Student Model}.

\begin{table*}[!hbtp]
    \centering
    \caption{Student model parameters applied}\label{Student Model}
    \begin{tabular*}{\textwidth}{@{\extracolsep\fill}llllll}
    \toprule
    Name & Kernel size/stride & Input size & Output size & Activation function & \#Parameters \\ 
    \midrule
    Conv1D & (64, ) / 8 & 1 $\times$ 1024 & 4 $\times$ 128 & ReLU & 260 \\
    Pooling & 2 / 2 & 4 $\times$ 128 & 4 $\times$ 64 &  & 0 \\
    FC & & 256 & 10 & & 2570 \\
    \multicolumn{5}{c}{Total of trainable parameters} & 2830 \\
    \hline
    \end{tabular*}
\end{table*}

The DKDL-Net model is a single layer network and framework shown in Fig. \ref{DKD model}, adding LoRA module before convolutional and fully connected layers, and the relevant parameter configurations for the model's input, output, and convolutional kernel size are presented in Table \ref{LoRA Student}.

\begin{table*}[!hbtp]
    \centering
    \caption{DKDL-Net model parameters applied}\label{LoRA Student}
    \begin{tabular*}{\textwidth}{@{\extracolsep\fill}llllll}
    \toprule
    Name & Kernel size/stride & Input & Output & Activation function & \#Parameters \\ 
    \midrule
    Conv1D\_LoRA & & 1 $\times$ 1024 & 4 $\times$ 128 &  & 816 \\
    Conv1D & (64, ) / 8 & 1 $\times$ 1024 & 4 $\times$ 128 & ReLU & 260 \\
    Pooling & 2 / 2 & 4 $\times$ 128 & 4 $\times$ 64 &  & 0 \\
    FC\_LoRA & & 256 & 10 & & 3192 \\
    FC & & 10 & 10 & & 2570 \\
    \multicolumn{5}{c}{Total of trainable parameters} & 6838 \\
    \hline
    \end{tabular*}
\end{table*}

\section{Experiments}\label{experiments}
\subsection{Experimental Configurations} \label{Experimental Configurations}
\textbf{Environment configuration.} All experiments for this model were conducted on a Windows 11 system with an Intel Core i7-9850H CPU at 2.60GHz and an NVIDIA GeForce GTX 1650 with Max-Q Design 4GB GPU. The code was run in an environment with Python 3.10.13 and PyTorch 2.0.1+cu117.

During the training of DKDL-Net, we utilized the Adaptive Moment Estimation (Adam) optimizer with a learning rate (LR) of 0.005 and a weight decay coefficient of 0.0001. We employed the cross-entropy loss function evaluate the loss between true and predicted labels.

\textbf{Baseline.} We selected MCNN-LSTM, FaultNet, BearingPGA-Net, KDSCNN, and WDCNN models as our baseline models. Among these, BearingPGA-Net and WDCNN are SOTA (state-of-the-art) models. However, BearingPGA-Net is more of a lightweight model, whereas WDCNN is a relatively large-scale model. KDSCNN is also a lightweight model and is comparable to our model.

\begin{figure}[!hbtp]
    \centering
    \includegraphics[width=0.6\linewidth]{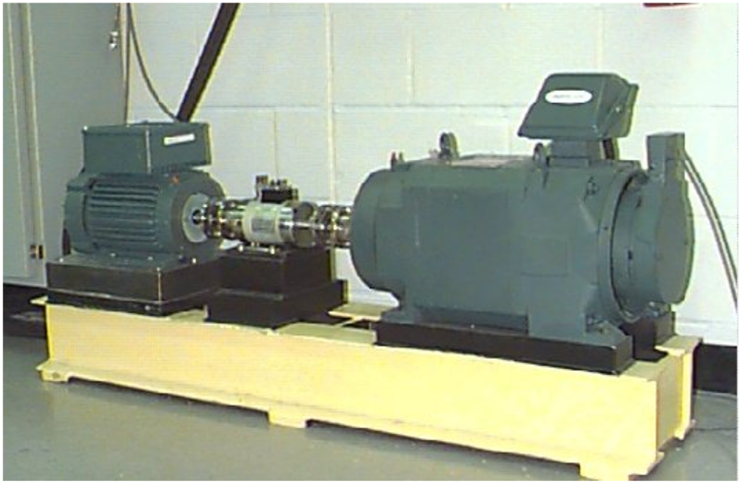}
    \caption{Data collection machine tools}
    \label{cwru_mechine}
\end{figure}

\textbf{Benchmark.} Our benchmark is based on the CWRU dataset, curated by the Case Western Reserve University Bearing Data Center. The machine to generate the CWRU dataset is shown in Fig. \ref{cwru_mechine}. This dataset includes ten categories, comprising nine types of faulty bearings and one healthy bearing. Vibration data are collected at 12 kHz and 48 kHz. The fault types and labels of the CWRU dataset are shown in Table \ref{CWRU}.

\begin{table*}[!hbtp]
    \caption{
The labels in the CWRU dataset and their corresponding fault types.}\label{CWRU}
    \begin{tabular*}{\textwidth}{@{\extracolsep\fill}llll}
    \toprule
    Faulty Mode & Fault size(mm) & Total dataset & class labels \\
    \midrule
    Health &  - & 280 & 0 \\
    Ball cracking (Minor) & 0.18 & 280 & 1   \\
    Ball cracking (Moderate) & 0.36 & 280 & 2 \\
    Ball cracking (Severe) & 0.53 & 280 & 3\\
    OR cracking (Minor) & 0.18 &    280 & 4 \\
    OR (Moderate) & 0.36 & 280 & 5 \\
    OR (Severe) & 0.53 & 280 & 6 \\
    IR (Minor) & 0.18 & 280 & 7 \\
    IR (Moderate) & 0.36 & 280 & 8 \\
    IR (Severe) & 0.53 & 280 & 9\\
    \hline
    \end{tabular*}
\end{table*}

\textbf{Evaluation metrics.} We use Accuracy, Precision, Recall and F1-Score to evaluate the performance of our model, which is calculated as in Eq. \ref{Accuracy} \ref{Precision} \ref{Recall} \ref{F1-Score}.
\begin{equation}\label{Accuracy}
    Accuracy = \frac{TP + TN}{TP + FP + FN + TN}
\end{equation}
\begin{equation}\label{Precision}
    Precision = \frac{TP}{TP + FP}
\end{equation}
\begin{equation}\label{Recall}
    Recall = \frac{TP}{TP + FN}
\end{equation}
\begin{equation}\label{F1-Score}
    F1-Score = \frac{2 \times Precision \times Recall}{Precision + Recall}
\end{equation}
where $TP$ represents True Positive, $TN$ represents True Negative, $FP$ represents False Positive, $FN$ represents False Negative.

\subsection{Bearing Fault Detection Experimental Results}

We conducted experiments on the CWRU dataset, and as shown in Table \ref{Model Compare}, the F1-Score of the DKDL-Net model is higher than that of the MCNN-LSTM, FaultNet, BearingPGA-Net, KDSCNN, and WDCNN models. Additionally, compared to the state-of-the-art (SOTA) model, our model achieved an improvement of 0.58\%. Despite having only 6838 parameters, our model has a higher accuracy than BearingPGA-Net by 0.58\%, with only 4008 more parameters. Compared to the KDSCNN model, our model has 948 more trainable parameters, yet it outperforms KDSCNN by 0.98\%. In summary, the DKDL-Net (our) model achieves higher accuracy while maintaining fewer parameters.

\begin{table}[!hbtp]
    \centering
    \caption{Comparison of bearing fault diagnosis F1-Score and parameters with different algorithms}\label{Model Compare}
    \begin{tabular*}{\textwidth}{@{\extracolsep\fill}lll}
    \toprule
    Model & F1-Score(\%) & Parameters \\
    \midrule
    MCNN-LSTM & 98.46 & 73480\\
    FaultNet & 98.50 & 627050\\
    BearingPGA-Net & 98.90 & \textbf{2830}\\
    KDSCNN &  98.50 & 5890\\
    WDCNN & 98.39 & 66790\\
    \textbf{DKDL-Net(our)} & \textbf{99.48} & 6838\\
    \bottomrule
    \end{tabular*}
\end{table}

As shown in Table \ref{Model Compare 3}, we evaluated the F1-Score, Precision, and Recall of the DKDL-Net model on the test dataset of CWRU. The DKDL-Net model outperforms the BearingPGA-Net, FaultNet, and MCNN-LSTM models in all three metrics. Compared to the best-performing BearingPGA-Net (SOTA) model, we achieved an improvement of nearly 0.55\% across all three metrics. In conclusion, our model is the best-performing model on the CWRU dataset.

\begin{table*}[!hbtp]
    \centering
    \caption{Comparison of bearing fault diagnosis Precision, Recall and F1-Score with different algorithms}\label{Model Compare 3}
    \begin{tabular*}{\textwidth}{@{\extracolsep\fill}llll}
    \toprule
    Model & Precision(\%) & Recall(\%) & F1-score(\%) \\
    \midrule
    MCNN-LSTM & 98.46 & 97.85 & 97.856 \\
    FaultNet & 98.60 & 98.57 & 98.57 \\
    BearingPGA-Net & 98.98 & 98.92 & 98.90 \\
    \textbf{DKDL-Net(our)} & \textbf{99.48} & \textbf{99.48} & \textbf{99.50} \\
    \hline
    \end{tabular*}
\end{table*}

As shown in Table \ref{Compare Model with S T DKDL-Net}, under the same configuration, our student model trained using Decoupled Knowledge Distillation (DKD) has 2,830 parameters. Compared to the teacher model, the student model's trainable parameters are reduced by approximately 95.93\%, but its F1-Score, Precision, and Recall decrease by 2.07\%, 1.92\%, and 2.08\%, respectively. This indicates that while the DKD model can compress model parameters, its accuracy significantly decreases.

\begin{table*}[!hbtp]
    \caption{Comparison of the parameters and assessment metrics of the teacher model, student model and DKDL-Net model}
    \label{Compare Model with S T DKDL-Net}
    \begin{tabular*}{\textwidth}{@{\extracolsep\fill}lllll}
    \toprule
    Model & Precision(\%) & Recall(\%) & F1-score(\%) & \#Parameters \\
    \midrule
    Teacher & \textbf{99.60} & \textbf{99.60} & \textbf{99.59} & 69626 \\
    Sudent & 97.68 & 97.52 & 97.52 & \textbf{2830} \\
    \textbf{DKDL-Net(our)} & 99.48 & 99.48 & 99.50 & 6838\\
    \hline
    \end{tabular*}
\end{table*}

On the other hand, the DKDL-Net model, based on DKD compression and LoRA fine-tuning, has 6,838 trainable parameters. Compared to the teacher model, the DKDL-Net model's trainable parameters are reduced by approximately 90.20\%, indicating that DKDL-Net can significantly reduce model complexity and resource requirements. The F1-Score, Precision, and Recall decrease by only 0.09\%, 0.12\%, and 0.12\%, respectively. This demonstrates that the accuracy loss caused by the DKDL-Net model compared to the teacher model is negligible. Therefore, the DKDL-Net model effectively compresses parameters while maintaining high accuracy.

In summary, our model achieves a compression ratio of 90.20\% with a negligible decrease in accuracy compared to the Teacher model, making it highly efficient in terms of compression while maintaining high accuracy.

\begin{figure}
    \centering
    \begin{subfigure}{0.32\linewidth}
        \centering
        \includegraphics[width=1\linewidth]{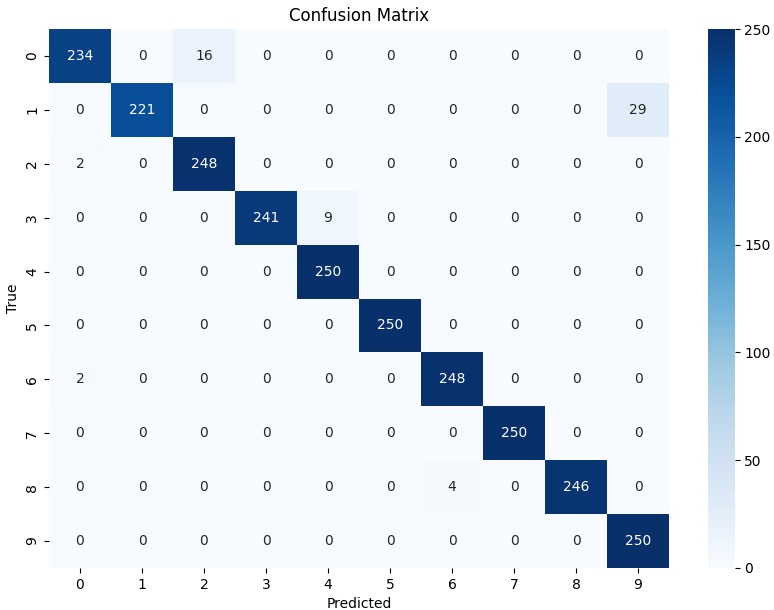}
        \caption{Student }
        \label{s confusion matrix}
    \end{subfigure}
    \begin{subfigure}{0.32\linewidth}
        \centering
        \includegraphics[width=1\linewidth]{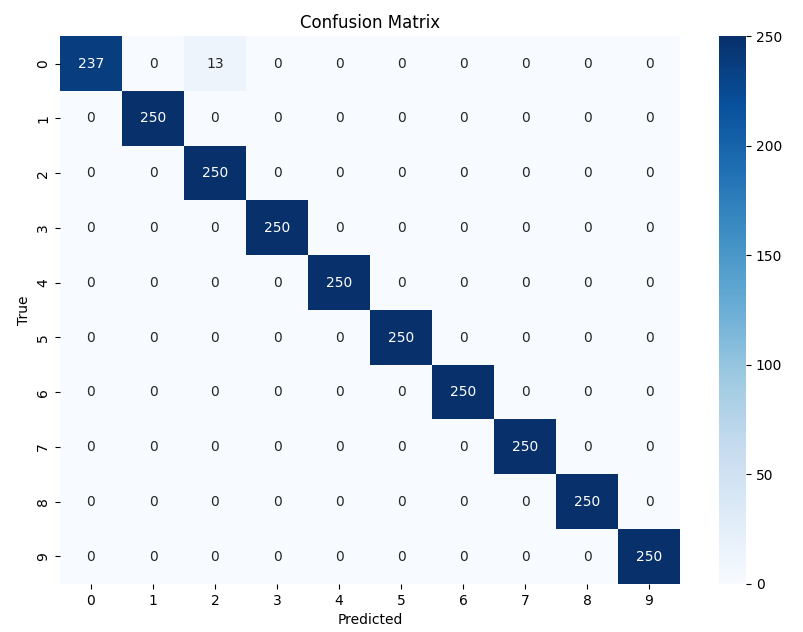}
        \caption{DKDL-Net}
        \label{DKDL-Net confusion matrix}
    \end{subfigure}
    \begin{subfigure}{0.32\linewidth}
        \centering
        \includegraphics[width=1\linewidth]{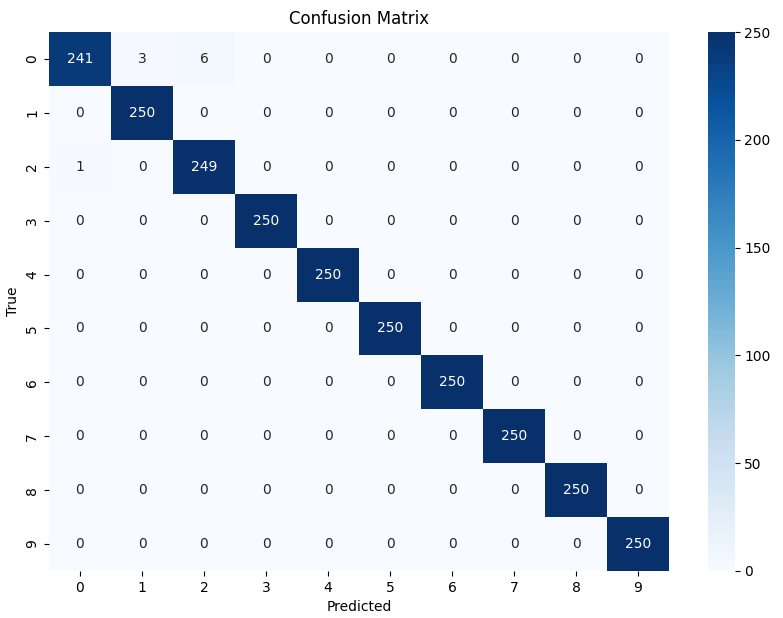}
        \caption{Teacher}
        \label{t confusion matrix}
    \end{subfigure}    
    \caption{Confusion Matrix for Student Model and DKDL-Net Model on CWRU Dataset}
    \label{DKDL-Net S T confusion matrix}
\end{figure}

Finally, we computed the confusion matrices for the DKDL-Net model as shown in Fig. \ref{DKDL-Net confusion matrix}, Student model as shown in Fig. \ref{s confusion matrix}, and Teacher model as shown in Fig. \ref{t confusion matrix} on the CWRU dataset, with 2,500 test samples, 250 samples per class.

\begin{figure}
    \centering
    \begin{subfigure}{0.49\linewidth}
        \centering
        \includegraphics[width=\linewidth]{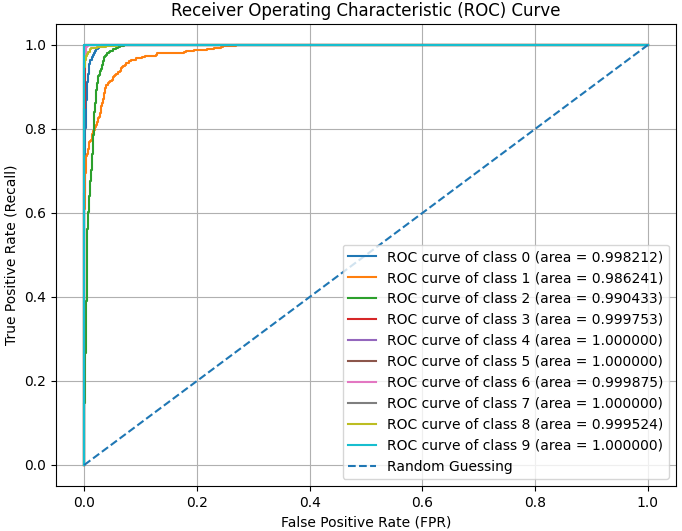}
        \caption{Student Model}
        \label{subfig:ROC_Student}
    \end{subfigure}
    \begin{subfigure}{0.49\linewidth}
        \centering
        \includegraphics[width=\linewidth]{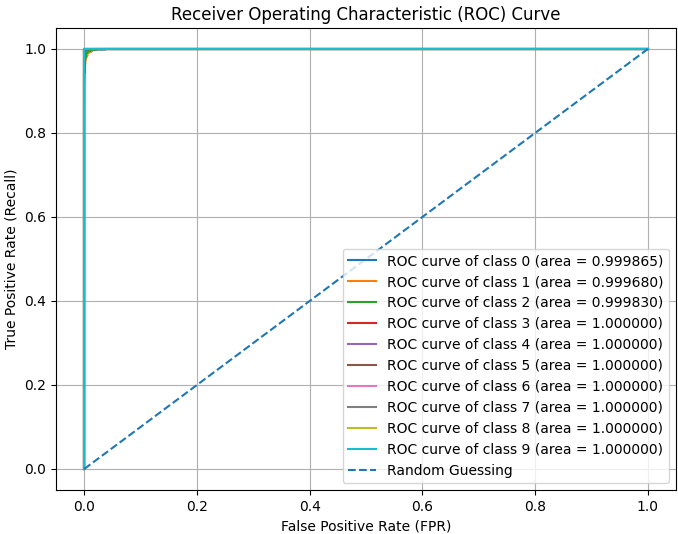}
        \caption{DKDL-Net Model}
        \label{subfig:ROC_DKDL-Net}
    \end{subfigure}
    \caption{ROC curves for student and DKDL-Net models on the CWRU dataset}
    \label{fig:ROC_both}
\end{figure}

We plotted the ROC curves for both the DKDL-Net model and the student model show in Fig. \ref{fig:ROC_both}. As shown in Fig. \ref{subfig:ROC_DKDL-Net} for the DKDL-Net model and Fig. \ref{subfig:ROC_Student} for the student model, it can be observed that the ROC curve for the student model is more jagged, while the ROC curve for the DKDL-Net model is smoother. Additionally, the area under the curve (AUC) for each class in the DKDL-Net model is generally larger than that of the student model. This indicates that our algorithm outperforms the student model's algorithm, demonstrating better performance.

We tested the DKDL-Net model on the CWRU bearing fault detection dataset. Under the same training configuration as see Section \ref{Experimental Configurations}, we tested 2500 samples, result as Shown in Table \ref{Inference}, and the DKDL-Net model required an average of 1757 $\mu$s per sample. DKDL-Net model with teacher model have 1x faster inference. This figure also demonstrates the high efficiency of our model.
\begin{table}[!hbtp]
    \centering
    \caption{Inference time for the DKDL-Net model}\label{Inference}
    \begin{tabular*}{\textwidth}{@{\extracolsep\fill}lll}
    \toprule
    Model & Num. of test samples & Avg. inference time($\mu$s
)\\
    \midrule
    Teacher & 2500 & 3816 \\
    \textbf{DKDL-Net (Our)} & 2500 & 1757 \\
    \bottomrule
    \end{tabular*}
\end{table}

Overall, the DKDL-Net model outperformed the Student model, and its performance was comparable to that of the Teacher model. Therefore, our DKDL-Net model can maintain good results even under high compression.

\section{Conclusion}\label{conclusion}
In this article, we propose a CNN model named DKDL-Net, which is based on decoupled knowledge distillation training and Low-Rank Adaptation fine-tuning. DKDL-Net is a single-layer neural network with only 6,838 parameters and an inference speed of 1,767 $\mu$s. It achieved an F1-Score of 99.50\% on the CRWU dataset, representing a 0.60\% improvement in F1-Score compared to the state-of-the-art (SOTA) models. Therefore, our model is highly efficient in detection while maintaining high accuracy. Moreover, the model is extremely lightweight, making it suitable for practical industrial applications.

\section{Compliance with ethical standards}

\textbf{Conflict of interest} The authors declare that they have no conflict of interest.

\bibliographystyle{unsrt}  
\bibliography{references}

\end{document}